\documentclass[11pt,a4paper]{article}

\usepackage{etoolbox}
\makeatletter
\patchcmd\@combinedblfloats{\box\@outputbox}{\unvbox\@outputbox}{}{\errmessage{\noexpand patch failed}}
\makeatother

\usepackage{times}
\usepackage{latexsym}
\usepackage{paralist}
\usepackage{subfig} 
\usepackage{placeins} 
\usepackage{algorithm}
\usepackage{booktabs} 
\usepackage{multirow}
\usepackage{graphicx}
\usepackage{amsmath}
\usepackage{siunitx}   
\usepackage{lipsum}
\usepackage{mathtools}
\usepackage{soul}
\setstcolor{red}

\newcommand*{\figref}[1]{\figurename~\ref{#1}}
\newcommand*{\tabref}[1]{\tablename~\ref{#1}}

\DeclareMathOperator*{\argmax}{arg\,max}
\newcommand\boldgray[1]{\textcolor{gray}{\textbf{#1}}}

\usepackage{url}
\usepackage[hyperref]{conll2018}

\aclfinalcopy 

\title{Simple Unsupervised Keyphrase Extraction using Sentence Embeddings}

\author{Kamil Bennani-Smires\textsuperscript{1}, Claudiu Musat\textsuperscript{1},  Andreaa Hossmann\textsuperscript{1},\\ \bf{Michael Baeriswyl\textsuperscript{1}, Martin Jaggi\textsuperscript{2}}  \\
    \textsuperscript{1}Data, Analytics \& AI, Swisscom AG\\
    \tt firstname.lastname@swisscom.com
    \\
    \textsuperscript{2}Machine Learning and Optimization Laboratory, EPFL \\
    \tt martin.jaggi@epfl.ch
}

\date{}

\begin{document}
\maketitle
\begin{abstract}
Keyphrase extraction is the task of automatically selecting a small set of phrases that best describe a given free text document.
Supervised keyphrase extraction requires large amounts of labeled training data and generalizes very poorly outside the domain of the training data. At the same time, unsupervised systems have poor accuracy, and often do not generalize well, as they require the input document to belong to a larger corpus also given as input.
Addressing these drawbacks, in this paper, we tackle keyphrase extraction from single documents with EmbedRank: a novel unsupervised method, that leverages sentence embeddings.
EmbedRank achieves higher F-scores than graph-based state of the art systems on standard datasets and is suitable for real-time processing of large amounts of Web data. With EmbedRank, we also explicitly increase coverage and diversity among the selected keyphrases by introducing an embedding-based maximal marginal relevance (MMR) for new phrases. A user study including over 200 votes showed that, although reducing the phrases' semantic overlap leads to no gains in F-score, our high diversity selection is preferred by humans.
\end{abstract}

\section{Introduction}
\label{sec:intro}
Document keywords and keyphrases enable faster and more accurate search in large text collections, serve as condensed document summaries, and are used for various other applications, such as categorization of documents. In particular, keyphrase extraction is a crucial component when gleaning real-time insights from large amounts of Web and social media data.
In this case, the extraction must be \emph{fast} and the keyphrases must be \emph{disjoint}. Most existing systems are slow and plagued by over-generation, i.e. extracting redundant keyphrases. 
Here, we address both these problems with a new unsupervised algorithm.

Unsupervised keyphrase extraction has a series of advantages over supervised methods. Supervised keyphrase extraction always requires the existence of a (large) annotated corpus of both documents and their manually selected keyphrases to train on - a very strong requirement in most cases. Supervised methods also perform poorly outside of the domain represented by the training corpus - a big issue, considering that the domain of new documents may not be known at all. Unsupervised keyphrase extraction addresses such information-constrained situations in one of two ways: \begin{inparaenum}[(a)]\item by relying on in-corpus statistical information (e.g., the inverse document frequency of the words), and the current document; \item by only using information extracted from the current document\end{inparaenum}.

We propose EmbedRank - an unsupervised method to automatically extract keyphrases from a document, that is both simple and \textit{only requires the current document itself}, rather than an entire corpus that this document may be linked to. Our method relies on notable new developments in text representation learning~\cite{Le2014,Kiros2015,Pagliardini2017}, where documents or word sequences of arbitrary length are embedded into the same continuous vector space. This opens the way to computing semantic relatedness among text fragments by using the induced similarity measures in that feature space. Using these semantic text representations, we guarantee the two most challenging properties of keyphrases: \emph{informativeness} obtained by the distance between the embedding of a candidate phrase and that of the full document; \emph{diversity} expressed by the distances among candidate phrases themselves.

In a traditional F-score evaluation, EmbedRank clearly \textbf{outperforms the current state of the art} (i.e. complex graph-based methods \cite{Mihalcea2004,Wan2008a,RuiWangWeiLiu2015}) 
on two out of three common datasets for keyphrase extraction. We also evaluated the impact of \textbf{ensuring diversity} by conducting a user study, since this aspect cannot be captured by the F-score evaluation. The study showed that users highly prefer keyphrases with the diversity property.
Finally, to the best of our knowledge, we are the first to present an unsupervised method based on phrase and document embeddings for keyphrase extraction, as opposed to standard individual word embeddings.

The paper is organized as follows. Related work on keyphrase extraction and sentence embeddings is presented in Section~\ref{sec:related}. In Section~\ref{sec:embrank} we present how our method works. An enhancement of the method allowing us to gain a control over the redundancy of the extracted keyphrases is then described in Section~\ref{sec:embmmr}. Section~\ref{section:expandresult} contains the different experiments that we performed and Section \ref{sec:discussion} outlines the importance of Embedrank in real-world examples.

\section{Related Work}
\label{sec:related}
A comprehensive, albeit slightly dated survey on keyphrase extraction is available~\cite{Hasan2011}. Here, we focus on unsupervised methods, as they are superior in many ways (domain independence, no training data) and represent the state of the art in performance. As EmbedRank relies heavily on (sentence) embeddings, we also discuss the state of the art in this area.

\subsection{Unsupervised Keyphrase Extraction}

Unsupervised keyphrase extraction comes in two flavors: corpus-dependent~\cite{Wan2008a} and corpus-independent.

Corpus-independent methods, including our proposed method, require no other inputs than the one document from which to extract keyphrases. Most such existing methods are graph-based, with the notable exceptions of KeyCluster~\cite{Liu2009} and TopicRank~\cite{Bougouin2013}. In graph-based keyphrase extraction, first introduced with TextRank~\cite{Mihalcea2004}, the target document is a graph, in which nodes represent words and edges represent the co-occurrence of the two endpoints inside some window. The edges may be weighted, like in SingleRank~\cite{Wan2008a}, using the number of co-occurrences as weights. The words (or nodes) are scored using some node ranking metric, such as degree centrality or PageRank~\cite{Page1998}. Scores of individual words are then aggregated into scores of multi-word phrases. Finally, sequences of consecutive words which respect a certain sequence of part-of-speech tags are considered as candidate phrases and ranked by their scores.
Recently, WordAttractionRank \cite{RuiWangWeiLiu2015} followed an approach similar to SingleRank, with the difference of using a new weighting scheme for edges between two words, to incorporate the distance between their word embedding representation. 
\newcite{florescu2017position} use node weights, favoring words appearing earlier in the text.

Scoring a candidate phrase as the aggregation of its words score \cite{Mihalcea2004,Wan2008a,florescu2017position} can lead to over-generation errors. This happens as several candidate phrases can obtain a high score because one of their consitutent words has a high score. This behavior leads to uninformative keyphrase with one important word present but lacking informativeness as a whole. In addition focusing on individual words hurts the diversity of the results.
\subsubsection{Diversifying results}
Ensuring diversity is important in the presentation of results to users in the information retrieval literature. Examples include MMR~\cite{Goldstein1998}, IA-Select~\cite{IAselect} or Max-Sum Diversification~\cite{maxsumdiv}. We used MMR in this work because of its simplicity in terms of both implementation and, more importantly, interpretation.\\ 
The following methods directly integrate a diversity factor in the way they are selecting keyphrases.
Departing from the popular graph approach, KeyCluster~\cite{Liu2009} introduces a clustering-based approach. The words present in the target document are clustered and, for each cluster, one word is selected as an ``exemplar term''. Candidate phrases are filtered as before, using the sequence of part-of-speech tags and, finally, candidates which contain at least one exemplar term are returned as the keyphrases.

TopicRank~\cite{Bougouin2013} combines the graph and clustering-based approaches. Candidate phrases are first clustered, then a graph where each node represents a cluster is created.
TopicRank clusters phrases based on the percentage of shared words, resulting in e.g., ``\textit{fantastic teacher}'' and ``\textit{great instructor}'' not being clustered together, despite expressing the same idea.
In the follow-up work using multipartite graphs \cite{multipartite}, the authors encode topical information within a multipartite graph structure.

In contrast, EmbedRank represents both the document and candidate phrases as vectors in a high-dimensional space, leveraging novel semantic document embedding methods beyond simple averaging of word vectors. In the resulting vector space, we can thus compute meaningful distances between a candidate phrase and the document (for informativeness), as well as the semantic distance between candidates (for diversity).
\subsection{Word and Sentence Embeddings}
Word embeddings~\cite{Mikolov2013} marked a very impactful advancement in representing words as vectors in a continuous vector space. Representing words with vectors in moderate dimensions solves several major drawbacks of the classic bag-of-words representation, including the lack of semantic relatedness between words and the very high dimensionality (size of the vocabulary).
Different methods are needed for representing entire sentences or documents.
Skip-Thought~\cite{Kiros2015} provides sentence embeddings trained to predict neighboring sentences. Paragraph Vector~\cite{Le2014} finds paragraph embeddings using an unordered list of paragraphs.
The method can be generalized to also work on sentences or entire documents, turning paragraph vectors into more generic document vectors~\cite{lau2016empirical}.

Sent2Vec~\cite{Pagliardini2017} uses word n-gram features to produce sentence embeddings.
It produces word and n-gram vectors specifically trained to be additively combined into a sentence vector, as opposed to general word-vectors. Sent2Vec 
features much faster inference than Paragraph Vector~\cite{Le2014} or Skip-Thought~\cite{Kiros2015}. Similarly to recent word and document embeddings, Sent2Vec reflects semantic relatedness between phrases when using standard similarity measures on the corresponding vectors. This property is at the core of our method, as we show it outperforms competing embedding methods for keyphrase extraction.

\section{EmbedRank: From Embeddings to Keyphrases}
\label{sec:embrank}

In this and the next section, we introduce and describe our novel keyphrase extraction method, EmbedRank~\footnote{\url{https://github.com/swisscom/ai-research-keyphrase-extraction}}. The method consists of three main steps, as follows:
\begin{inparaenum}[(1)]
\item We extract candidate phrases from the text, based on part-of-speech sequences. More precisely, we keep only those phrases that consist of zero or more adjectives followed by one or multiple nouns~\cite{Wan2008a}.
\item We use sentence embeddings to represent (\textbf{embed}), both the candidate phrases and the document itself in the same high-dimensional vector space (Sec. \ref{sec:embed}).
\item We \textbf{rank} the candidate phrases to select the output keyphrases (Sec. \ref{sec:rank}).
\end{inparaenum}
In addition, in the next section, we show how to improve the ranking step, by providing a way to tune the diversity of the extracted keyphrases.

\begin{figure*}
\subfloat[EmbedRank (without diversity)]{\hspace{-0.7cm}\includegraphics[width=0.57\textwidth]{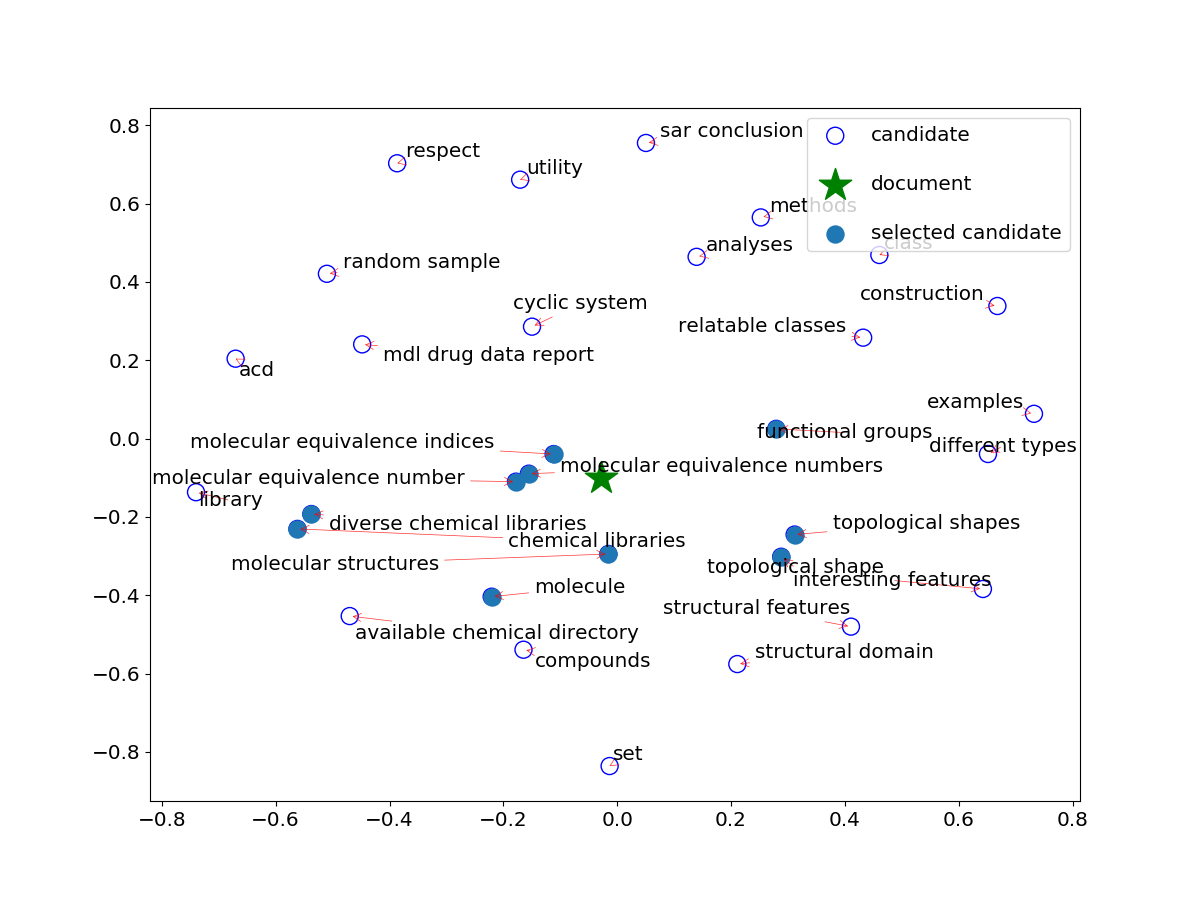}\vspace{-0.2cm}\label{fig:embeddspace_big}}\hspace{-1.3cm}
\subfloat[EmbedRank++ (with diversity)]{\includegraphics[width=0.57\textwidth]{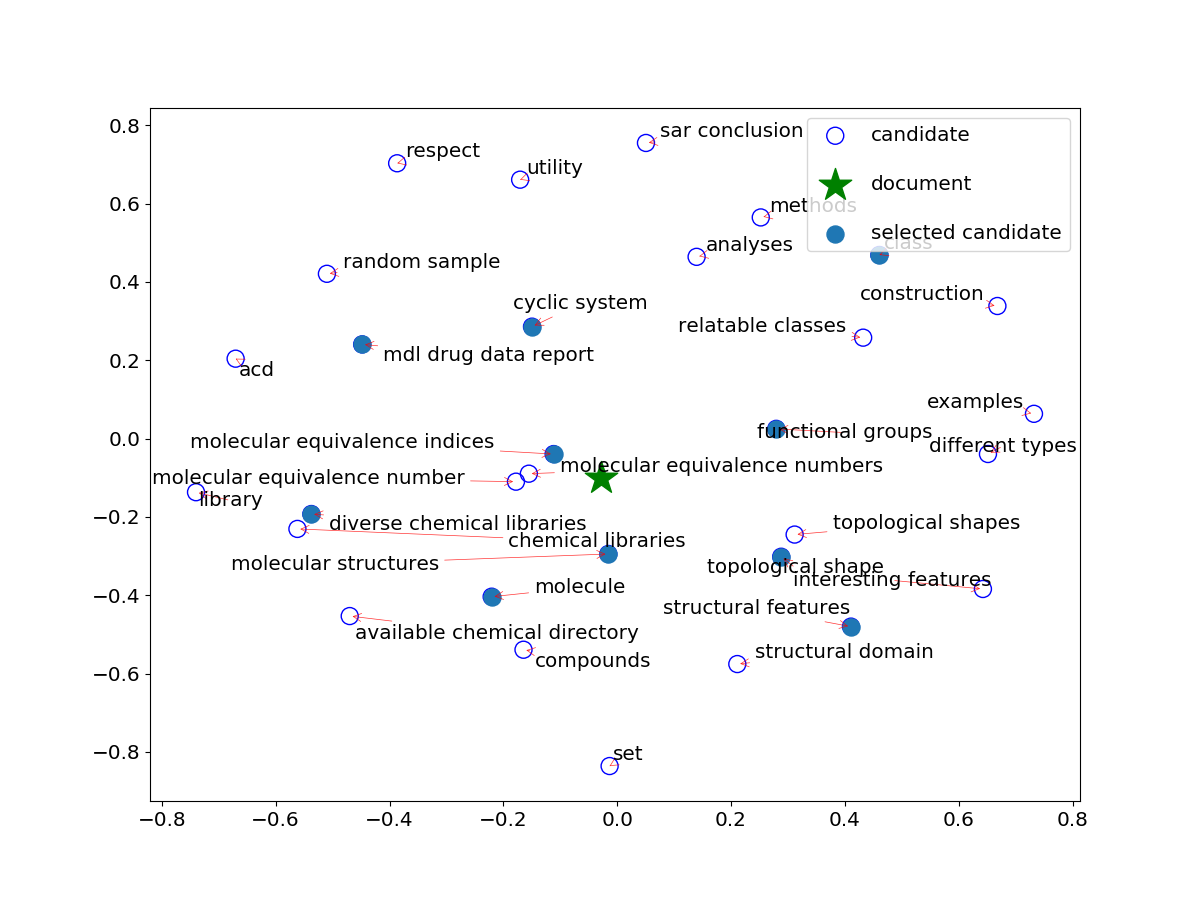}\vspace{-0.2cm}\label{fig:embeddspace05}}
\caption[]{Embedding space\footnotemark~~of a scientific abstract entitled \textit{``Using molecular equivalence numbers to visually explore structural features that distinguish chemical libraries''}}
\vspace{-0.2cm}
\label{fig:embeddspace}
\end{figure*}

\subsection{Embedding the Phrases and the Document}
\label{sec:embed}
State-of-the-art text embeddings (word, sentence, document) capture semantic relatedness via the distances between the corresponding vector representations within the shared vector space. We use this property to rank the candidate phrases extracted in the previous step, by measuring their distance to the original document. Thus, semantic relatedness between a candidate phrase and its document becomes a proxy for informativeness of the phrase.

Concretely, this second step of our keyphrase extraction method consists of:
\begin{enumerate}[(a)]
\item Computing the \textit{document embedding}. This includes a noise reduction procedure, where we keep only the adjectives and nouns contained in the input document.
\item Computing the \textit{embedding of each candidate phrase} separately, again with the same algorithm.
\end{enumerate}

To determine the impact the document embedding method may have on the final outcome, we evaluate keyphrases obtained using both the popular Doc2Vec~\cite{lau2016empirical} (denoted EmbedRank d2v) and ones based on the newer Sent2vec~\cite{Pagliardini2017} (denoted EmbedRank s2v).
\footnotetext{Visualization based on multidimensional scaling with cosine distance on the original $Z=Z_s=700$ dimensional embeddings.}
Both embedding methods \textbf{allow us to embed arbitrary-length sequences of words}. To embed both phrases and documents, we employ publicly available pre-trained models of Sent2Vec\footnote{\url{https://github.com/epfml/sent2vec}} and Doc2vec\footnote{\url{https://github.com/jhlau/doc2vec}}. The pre-computed Sent2vec embeddings based on words and n-grams vectors have $Z = Z_s=700$ dimensions, while for Doc2vec $Z = Z_d=300$. All embeddings are trained on the large English Wikipedia corpus.\footnote{The generality of this corpus, as well as the unsupervised embedding method itself ensure that the computed text representations are general-purpose, thus domain-independent.} 
EmbedRank s2v is very fast, since Sent2vec infers a document embedding from the pre-trained model, by averaging the pre-computed representations of the text's components (words and n-grams), in a single linear pass through the text. EmbedRank d2v is slower, as Doc2vec uses the embedding network to infer a vector for the whole document. Both methods provide vectors comparable in the same semantic space, no matter if the input "document" is a word, a phrase, a sentence or an entire document.

After this step, we have one $Z$-dimensional vector representing our document and a $Z$-dimensional vector for each of our candidate phrases, all sharing the same reference space. Figure~\ref{fig:embeddspace} shows a concrete example, using EmbedRank s2v, from one of the datasets we used for evaluation (scientific abstracts). As can be seen by comparing document titles and candidate phrases, our initial assumption holds in this example: the closer a phrase is to the document vector, the more informative that phrase is for the document. Therefore, it is sensible to use the cosine similarity between the embedding of the candidate phrase and the document embedding as a measure of informativeness.

\subsection{Selecting the Top Candidates}
\label{sec:rank}
Based on the above, we select the top keyphrases out of the initial set, by ranking the candidate phrases according to their cosine distance to the document embedding. In~\figref{fig:embeddspace}, this results in ten highlighted keyphrases, which are clearly in line with the document's title.

\begin{table*}
\resizebox{\textwidth}{!}{
\begin{tabular}{ccccccccc}
\toprule
Dataset & Documents & Avg tok & Avg cand & Keyphrases & Avg kp & Missing kp in doc & Missing kp in cand & Missing due to cand \\
\midrule
 \textbf{Inspec} &       500 &        134.63 &              26.39 &       4903 &               9.81 &                    21.52\% &                           39.85\% &                             18.34\% \\
    \textbf{DUC} &       308 &        850.02 &             138.47 &       2479 &               8.05 &                     2.18\% &                           12.38\% &                             10.21\% \\
 \textbf{NUS} &       209 &       8448.55 &             765.56 &       2272 &              10.87 &                    14.39\% &                           30.85\% &                             16.46\% \\
\bottomrule
\end{tabular}}
 \caption{The three datasets we use. Columns are: number of documents; average number of tokens per document; average number of unique candidates per document; total number of unique keyphrases; average number of unique keyphrases per document; percentage of keyphrases not present in the documents; percentage of keyphrases not present in the candidates; percentage of keyphrases present in the document, but not in the candidates. These statistics were computed after stemming the candidates, the keyphrases and the document.}
 \label{table:datasetstat}
\vspace{-0.2cm}
\end{table*}

Nevertheless, it is notable that there can be significant redundancy in the set of top keyphrases. For example, ``\textit{molecular equivalence numbers}'' and ``\textit{molecular equivalence indices}'' are both selected as separate keyphrases, despite expressing the same meaning. This problem can be elegantly solved by once again using our phrase embeddings and their cosine similarity as a proxy for semantic relatedness. We describe our proposed solution to this in the next section.

Summarizing this section, we have proposed an unsupervised step-by-step method to extract \textit{informative keyphrases} from a single document by using sentence embeddings.

\section{EmbedRank++: Increasing Keyphrase Diversity with MMR}
\label{sec:embmmr}
By returning the $N$ candidate phrases closest to the document embedding, EmbedRank only accounts for the phrase informativeness property, leading to redundant keyphrases.
In scenarios where users directly see the extracted keyphrases (e.g. text summarization, tagging for search), this is problematic: redundant keyphrases adversely impact the user's experience. This can deteriorate to the point in which providing keyphrases becomes completely useless.

Moreover, if we extract a fixed number of top keyphrases, redundancy hinders the diversification of the extracted keyphrases. In the document from Figure \ref{fig:embeddspace}, the extracted keyphrases include \{\textit{topological shape, topological shapes}\} and \{\textit{molecular equivalence number, molecular equivalence numbers, molecular equivalence indices}\}. That is, four out of the ten keyphrase ``slots'' are taken by redundant phrases.

This resembles search result diversification~\cite{Drosou2010}, where a search engine balance query-document relevance and document diversity. One of the simplest and most effective solutions to this is the Maximal Marginal Relevance (MMR)~\cite{Goldstein1998} metric, which combines in a controllable way the concepts of relevance and diversity. We show how to adapt MMR to keyphrase extraction, in order to combine keyphrase informativeness with dissimilarity among selected keyphrases.

The original MMR from information retrieval and text summarization is based on the set of all initially retrieved documents, $R$, for a given input query $Q$, and on an initially empty set $S$ representing documents that are selected as good answers for $Q$. $S$ is iteratively populated by computing MMR as described in \eqref{eq:originalmmr}, where $D_i$ and $D_j$ are retrieved documents, and $Sim_1$ and $Sim_2$ are similarity functions.
\begin{equation}
\label{eq:originalmmr}
\begin{split}
\text{MMR}  :=  \argmax_{D_i \in R\setminus S} \bigg{[} \lambda\cdot Sim_1(D_i,Q) \\
  - (1-\lambda)\max_{D_j \in S} Sim_2(D_i, D_j)  \bigg{]}
\end{split}
\end{equation}

When $\lambda = 1$ MMR computes a standard, relevance-ranked list, while when $\lambda = 0$ it computes a maximal diversity ranking of the documents in $R$.
To use MMR here, we adapt the original equation as:
\begin{equation}
\label{eq:kpmmr}
\begin{split}
\text{MMR} :=  \argmax_{C_i \in C\setminus K}\bigg[\lambda\cdot\widetilde{cos_{sim}}(C_i,\text{doc}) \\
               - (1-\lambda)\max_{C_j \in K} \widetilde{cos_{sim}}(C_i, C_j)\bigg],
\end{split}
\end{equation}
where $C$ is the set of candidate keyphrases, $K$ is the set of extracted keyphrases, $doc$ is the full document embedding, $C_i$ and $C_j$ are the embeddings of candidate phrases $i$ and $j$, respectively. 
Finally, $\widetilde{cos_{sim}}$ is a normalized cosine similarity~\cite{Mori2003}, described by the following equations. This ensures that, when $\lambda = 0.5$, the relevance and diversity parts of the equation have equal importance.
\begin{subequations}
\begin{equation}
\label{eq:normdoc1}
\begin{split}
& \widetilde{cos_{sim}}(C_i, \text{doc}) =  0.5 + \\
& \frac{ncos_{sim}(C_i, \text{doc}) - \overline{ncos_{sim}(C, \text{doc})}}{\sigma(ncos_{sim}(C, \text{doc}))}.
\end{split}
\end{equation}
\begin{equation}
\label{eq:normdoc0}
\begin{split}
&ncos_{sim}(C_i, \text{doc}) = \\
&\frac{cos_{sim}(C_i, \text{doc})- \min\limits_{C_j \in C} cos_{sim}(C_j, \text{doc})}{\max\limits_{C_j \in C} cos_{sim}(C_j, \text{doc}) }
\end{split}
\end{equation}
\end{subequations}

We apply an analogous transformation for the similarities between candidate phrases.

Summarizing, the method in the previous section is equivalent to using MMR for keyphrase extraction from Equation~\eqref{eq:kpmmr} with $\lambda = 1$. 
The generalized version of the algorithm, EmbedRank++, remains the same, except for the last step, where we instead use Equation~\eqref{eq:kpmmr} to perform the final selection of the $N$ candidates, therefore returning simultaneously relevant and diverse keyphrases, tuned by the trade-off parameter $\lambda$.

\section{Experiments and results}
\label{section:expandresult}
\sisetup{detect-weight=true,detect-inline-weight=math}
\begin{table*}[!htb]
  \label{tab:commands}
  \resizebox{\linewidth}{!}{

 \begin{tabular}{clSSSSSSSSSSS}
    \toprule
\multirow{2}*{N}   &   \multirow{2}*{Method} &   \multicolumn{3}{c}    {Inspec}    & &  \multicolumn{3}{c}{DUC} && \multicolumn{3}{c}{NUS} \\

 &  &   {P} &   {R}  &   {F$_1$} &&   {P} &   {R} &   {F$_1$} && {P} & {R} & {F$_1$}  \\ \midrule

 \multirow{3}*{5} & TextRank & 24.87 & 10.46 & 14.72 & & 19.83 & 12.28 & 15.17 & & 5.00 & 2.36 & 3.21 \\
 & SingleRank & 38.18 & 23.26 & 28.91 & & 30.31 & 19.50 & 23.73 & & 4.06 & 1.90 & 2.58 \\
 & TopicRank & 33.25 & 19.94 & 24.93 && 27.80 & 18.28 & 22.05 && 16.94 &  ~~8.99 & 11.75 \\
  & Multipartite & 34.61 & 20.54 & 25.78 && 29.49 & 19.42 & 23.41 && \bfseries 19.23 & \bfseries 10.18 & \bfseries 13.31\\
 & WordAttractionRank & 38.55 & 23.55 & 29.24 && 30.83 & 19.79 & 24.11 &&  4.09 &  1.96 &  2.65 \\
 & EmbedRank d2v & \bfseries 41.49 &  \bfseries 25.40 & \bfseries 31.51 && 30.87 & 19.66 & 24.02 && 3.88 & 1.68 & 2.35 \\
 & EmbedRank s2v  & 39.63 &  23.98 &  29.88 & & \bfseries 34.84 & \bfseries 22.26 & \bfseries 27.16 && 5.53 & 2.44 & 3.39 \\
 &EmbedRank++ s2v ($\lambda=0.5$) & 37.44 & 22.28 &  27.94 && 24.75 & 16.20 &  19.58 & & 2.78 & 1.24 &  1.72 \\
 &EmbedRank$_{positional}$ s2v & 38.84 & 23.77 &  29.49 && \boldgray{39.53} & \boldgray{25.23} & \boldgray{30.80} & & 15.07 & 7.80 &  10.28 \\
 \midrule
 \multirow{3}*{10} & TextRank & 22.99 & 11.44 & 15.28 && 13.93 & 16.83 & 15.24 && 6.54 & 6.59 &  6.56 \\
 & SingleRank & 34.29 & 39.04 & 36.51 &&  24.74 & 30.97 & 27.51 && 5.22 & 5.04 & 5.13\\
 & TopicRank & 27.43 & 30.8 & 29.02 && 21.49 & 27.26 & 24.04 && 13.68 &  13.94 & 13.81 \\
  & Multipartite & 28.07 & 32.24 & 30.01 && 22.50 & 28.85 & 25.28 && \bfseries 16.51 & \bfseries 17.36 & \bfseries 16.92 \\
  & WordAttractionRank & 34.10 & 38.94 & 36.36 && 25.06 & 31.41 & 27.88 &&  5.15 &  5.12 &  5.14 \\
 & EmbedRank d2v & \bfseries 35.75 & \bfseries 40.40 & \bfseries 37.94 && 25.38 & 31.53 & 28.12 && 3.95 & 3.28 & 3.58 \\
 & EmbedRank s2v &  34.97 &  39.49 &  37.09 && \bfseries 28.82 & \bfseries 35.58 & \bfseries 31.85 && 5.69 & 5.18 & 5.42 \\
 &EmbedRank++ s2v ($\lambda=0.5$) & 30.31 & 34.29 &  32.18  && 18.27 & 23.34 &  20.50 && 1.91 & 1.69 &  1.79 \\
 &EmbedRank$_{positional}$ s2v & 32.46 & 36.61 &  34.41 && \boldgray{32.23} & \boldgray{39.95} & \boldgray{35.68} & & 13.50 & 13.36 &  13.43 \\
  \midrule
 \multirow{3}*{15} & TextRank &  22.80 & 11.50 & 15.29 && 11.25 & 19.21 & 14.19 && 6.14 & 9.16 & 7.35 \\
 & SingleRank & 30.91 & 48.92 & 37.88 && 21.20 & 38.77 & 27.41 && 5.42 & 8.24 & 6.54 \\
 & TopicRank & 24.51 & 37.45 & 29.62 && 17.78 & 32.92 & 23.09 && 11.04 & 16.47 &  13.22 \\
  & Multipartite &  25.38 & 41.32 & 31.44 && 19.72 & 36.87 & 25.7 && \bfseries 14.13 & \bfseries 21.86 & \bfseries 17.16 \\
 & WordAttractionRank & 30.74 & 48.62 & 37.66 && 21.82 & 40.05 & 28.25 &&  5.11 &  7.41 &  6.05 \\ 
 & EmbedRank d2v & 31.06 & 48.80 & 37.96 && 22.37 & 40.48 & 28.82 && 4.33 & 5.89 & 4.99 \\
 & EmbedRank s2v  & \bfseries 31.48 & \bfseries 49.23 & \bfseries 38.40 && \bfseries 24.49 & \bfseries 44.20 & \bfseries 31.52 && 5.34 & 7.06 & 6.08 \\
 &EmbedRank++ s2v ($\lambda=0.5$) & 27.24 & 43.25 &  33.43 & & 14.86 & 27.64 &  19.33 && 1.59 & 2.06 &  1.80 \\
 &EmbedRank$_{positional}$ s2v & 29.44 & 46.25 &  35.98 && \boldgray{27.38} & \boldgray{49.73} & \boldgray{35.31} && 12.27 &  17.63 & 14.47 \\
    \bottomrule
  \end{tabular}
  }
\caption{Comparison of our method with state of the art on the three datasets. Precision (P), Recall (R), and F-score (F$_1$) at 5, 10, 15 are reported. Two variations of EmbedRank with $\lambda = 1$ are presented: s2v uses Sent2Vec embeddings, while d2v uses Doc2Vec. }
  \label{table:result}
\end{table*}

In this section we show that EmbedRank outperforms the graph-based state-of-the-art schemes on the most common datasets, when using traditional F-score evaluation. In addition, we report on the results of a sizable user study showing that, although EmbedRank++ achieves slightly lower F-scores than EmbedRank, users prefer the semantically diverse keyphrases it returns to those computed by the other method.

\subsection{Datasets}
\tabref{table:datasetstat} describes three common datasets for keyphrase extraction. 
\\
The \textbf{Inspec} dataset \cite{Hulth:2003:IAK:1119355.1119383} consists of 2\,000 short documents from scientific journal abstracts. To compare with previous work \cite{Mihalcea2004,Hasan2010,Bougouin2013,Wan2008a}, we evaluated our methods on the test dataset (500 documents).
\\
\textbf{DUC 2001} \cite{Wan2008a} consists of 308 medium length newspaper articles from TREC-9. The documents originate from several newspapers and are organized in 30 topics. For keyphrase extraction, we used exclusively the text contained in the first \textsf{$<$TEXT$>$} tags of the original documents (we do not use titles and other metadata).
\\
\textbf{NUS}~\cite{Nguyen2007} consists of 211 long documents (full scientific conference papers), of between 4 and 12 pages. Each document has several sets of keyphrases: one created by the authors and, potentially, several others created by annotators. Following \newcite{Hasan2010}, we evaluate on the union of all sets of assigned keyphrases (author and annotator(s)).
The dataset is very similar to the SemEval dataset which is also often used for keyphrase extraction. Since our results on SemEval are very similar to NUS, we omit them due to space constraints.

As shown in \tabref{table:datasetstat}, not all assigned keyphrases are present in the documents (missing kp in doc). It is thus impossible to achieve a recall of $100\%$. We show in the next subsection that our method beats the state of the art on short scientific documents and clearly outperforms it on medium length news articles.
 
\subsection{Performance Comparison}
We compare EmbedRank s2v and d2v (no diversity) to five state-of-the-art, corpus-independent methods\footnote{TextRank, SingleRank, WordAttractionRank were implemented using the graph-tool library~\url{https://graph-tool.skewed.de}. We reset the co-occurence window on new sentence.}: TextRank~\cite{Mihalcea2004}, SingleRank~\cite{Wan2008a}, WordAttractionRank~\cite{RuiWangWeiLiu2015}, TopicRank\footnote{\url{https://github.com/boudinfl/pke}}~\cite{Bougouin2013} and
Multipartite~\cite{multipartite}.

For TextRank and SingleRank, we set the window size to $2$ and to $10$ respectively, i.e. the values used in the respective papers. We used the same PoS tagged text for all methods. For both underlying d2v and s2v document embedding methods, we use their standard settings as described in Section \ref{sec:embrank}.
We followed the common practice to stem - with the Porter Stemmer~\cite{journals/program/Porter80} - the extracted and assigned keyphrases when computing the number of true positives.

As shown in \tabref{table:result}, EmbedRank outperforms competing methods on two of the three datasets in terms of precision, recall, and Macro F$_1$ score. In the context of typical Web-oriented use cases, most data comes as either very short documents (e.g. tweets) or medium ones (e.g. news articles). The expected performance for Web applications is thus closer to the one observed on the Inspec and DUC2001 datasets, rather than on NUS.

However, on long documents, Multipartite outperforms all other methods. The most plausible explanation is that Multipartite, like TopicRank incorporates positional information about the candidates. Using this feature leads to an important gain on long documents -- not using it can lead to a 90\% relative drop in F-score for TopicRank. We verify this intuition in the context of EmbedRank by naively multiplying the distance of a candidate to the document by the candidate's normalized offset position. 
We thus confirm the "positional bias" hypothesis, with EmbedRank$_{positional}$ matching the TopicRank scores on long documents and approaching the Multipartite ones.
The Multipartite results underline the importance of explicitly representing topics for long documents. This does not hold for short and medium documents, where the semantic information is successfully captured by the topology of the embedding space.

EmbedRank$_{positional}$ also outperforms on medium-length documents but, as the assumption that the keyphrases appear in a decreasing order of importance is very strong for the general case, we gray out the results, to stress the importance of the more generic EmbedRank variants.

The results also show that the choice of document embeddings has a high impact on the keyphrase quality. Compared to EmbedRank d2v, EmbedRank s2v is significantly better for DUC2001 and NUS, regardless of how many phrases are extracted. On Inspec however, changing the embeddings from doc2vec to sent2vec made almost no difference. A possible explanation is that, given the small size of the original text, the extracted keyphrases have a high likelihood of being single words, thus removing the advantage of having better embeddings for word groups. In all other cases, the results show a clear accuracy gain of Sent2Vec over Doc2Vec, adding to the practical advantage of improved inference speed for very large datasets.

\subsection{Keyphrase Diversity and Human Preference}
In this section, we add EmbedRank++ to the evaluation using the same three datasets. We fixed $\lambda$ to $0.5$ in the adapted MMR equation~\eqref{eq:kpmmr}, to ensure equal importance to informativeness and diversity.
As shown in Figure~\ref{fig:embeddspace05}, EmbedRank++ reduces the redundancy we faced with EmbedRank. 
However, EmbedRank++ surprisingly results in a decrease of the F-score, as shown in \tabref{table:result}.

We conducted a user study where we asked people to choose between two sets of extracted keyphrases: one generated with EmbedRank ($\lambda = 1$) and another with EmbedRank++ ($\lambda = 0.5$). We set $N$ to the number of assigned keyphrases for each document. During the study, we provided the annotators with the original text, and ask them to choose between the two sets.

For this user study, we randomly selected 20 documents from the Inspec and 20 documents from the DUC2001 dataset, collected 214 binary user preference votes. The long scientific papers (NUS) were included in the study, as the full papers were considered too long and too difficult for non-experts to comprehend and summarize.

\begin{figure}
\centering
\includegraphics[width=0.31\textwidth]{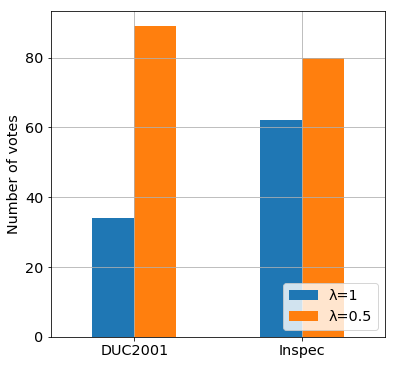}
\caption{User study among 20 documents from Inspec and 20 documents from DUC2001. Users were asked to choose their preferred set of keyphrases between the one extracted with EmbedRank++ ($\lambda = 0.5$) and the one extracted with EmbedRank ($\lambda = 1$).}
\label{fig:surveyresult}
\vspace{-0.2cm}
\end{figure}

As shown in \figref{fig:surveyresult}, users largely prefer the keyphrase extracted with EmbedRank++ ($\lambda = 0.5$). This is a major finding, as it is in contradiction with the F-scores given in \tabref{table:result}. If the result is confirmed by future tests, it casts a shadow on using solely F-score as an evaluation measure for keyphrase quality. A similar issue was shown to be present in Information Retrieval test collections \cite{DBLP:journals/ir/TononDC15}, and calls for research on new evaluation methodologies. We acknowledge that the presented study is a preliminary one and does not support a strong claim about the usefulness of the F-score for the given problem. It does however show that people dislike redundancy in summaries and that the $\lambda < 1$ parameter in EmbedRank is a promising way of reducing it.

Our intuition behind this novel result is that the EmbedRank method ($\lambda = 1$), as well as WordAttractionRank, SingleRank and TextRank can suffer from an accumulation of redundant keyphrases in which a true positive is present. By restricting the redundancy with EmbedRank++, we can select a keyphrase that is not present in the gold keyphrases, but expresses the same idea. The current F-score evaluation penalizes us as if we had chosen an unrelated keyphrase.

\section{Discussion}
\label{sec:discussion}
\begin{figure}
{\hspace{0cm}\includegraphics[width=0.5\textwidth]{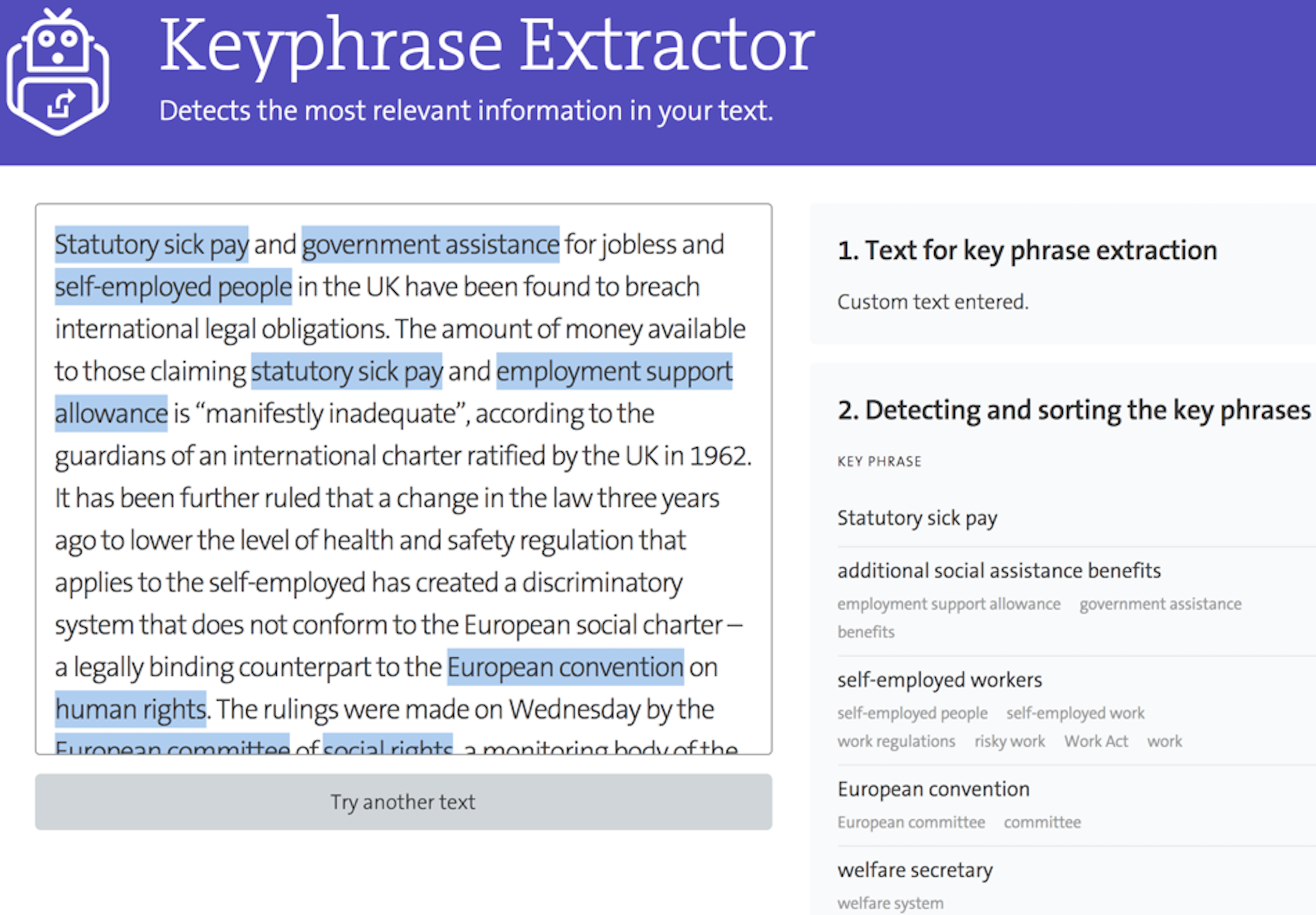}\vspace{-0.2cm}}\hspace{-1.3cm}\vspace{-2mm}
\caption[]{Keyphrase Grouping in news articles}
\label{fig:example}
\end{figure}

The usefulness of the corpus-free approach is in that we can extract keyphrases in any environment, for instance for news articles.
In Figure~\ref{fig:example} we show the keyphrases extracted from a sample article. The EmbedRank keyphrase extraction is fast, enabling real time computation and visualization. The disjoint nature of the EmbedRank keyphrases make them highly readable, creating a succinct summary of the original article.

By performing the analysis at phrase instead of word level, EmbedRank opens the possibility of grouping candidates with keyphrases before presenting them to the user. Phrases within a group have similar embeddings, like \textit{additional social assistance benefits, employment support allowance} and \textit{government assistance benefits}. Multiple strategies can be employed to select the most visible phrase - for instance the one with the highest score or the longest one. This grouping counters the over-generation problem.

\section{Conclusion}
\label{sec:conclusion}
In this paper we presented EmbedRank and EmbedRank++, two simple and scalable methods for keyphrase extraction from a single document, that leverage sentence embeddings. Both methods are entirely unsupervised, corpus-independent, and they only require the current document itself, rather than the entire corpus to which it belongs (that might not exist at all). They both depart from traditional methods for keyphrase extraction based on graph representations of the input text, and fully embrace sentence embeddings and their ability to model informativeness and diversity.

EmbedRank can be implemented on top of any underlying document embeddings, provided that these embeddings can encode documents of arbitrary length. We compared the results obtained with Doc2Vec and Sent2Vec, the latter one being much faster at inference time, which is important in a Web-scale setting. We showed that on short and medium length documents, EmbedRank based on Sent2Vec consistently improves the state of the art.
Additionally, thanks to a fairly large user study that we run, we showed that users appreciate diversity of keyphrases, and we raised questions on the reliability of evaluations of keyphrase extraction systems based on F-score.

\bibliography{simplekp}
\bibliographystyle{acl_natbib_nourl}
\end{document}